\documentclass{article} 
\usepackage{nips15submit_e,times}
\usepackage{hyperref}
\usepackage{url}
\usepackage{graphicx}
\usepackage{amsmath}
\usepackage{amssymb}
\usepackage{subfig}

\title{Can Boosting with SVM as Week Learners Help?}
\author{
Dinesh Govindaraj\\
Work Done: 2009 \\
Computer Science and Automation\\
Indian Institute of Science\\
Bangalore, India\\
}

\nipsfinalcopy 
\begin{document}
\maketitle

\begin{abstract}
Object recognition in images involves identifying objects with partial occlusions, viewpoint changes, varying illumination, cluttered backgrounds. Recent work in object recognition uses machine learning techniques SVM-KNN, Local Ensemble Kernel Learning, Multiple Kernel Learning. In this paper, we want to utilize SVM as week learners in AdaBoost. Experiments are done with classifiers like nearest neighbor, k-nearest neighbor, Support vector machines, Local learning(SVM-KNN) and AdaBoost. Models use Scale-Invariant descriptors and Pyramid histogram of gradient descriptors. AdaBoost is trained with set of week classifier as SVMs, each with kernel distance function on different descriptors. Results shows AdaBoost with SVM outperform other methods for Object Categorization dataset. 
\end{abstract}

\section{Survey of Recent Work}
SVM-KNN gets motivation from Local learning by Bottou and Vapnik uses K-Nearest neighbor to select local training point and uses SVM algorithm in those local training points for classification of object. Main problem here is time taken for classification. Local ensemble kernel learning was proposed to combine kernels locally. Multiple kernel learning for object recognition which uses PHOG feature vectors and shape descriptors gives state of art accuracy on CalTech-101. This method tries to learn discriminative power-invariance trade off.
\subsection{Local Learning and SVM-KNN}
Vapnik and Bottou \cite{zhang06} proposed a local learning algorithm for optical character recognition. They proposed simple algorithm: For each testing pattern, 
Select few training patterns in the vicinity of the testing pattern. 
Train a neural network with only these few examples.
Apply the resulting network to the testing pattern.
They proved that this simple algorithm will improve accuracy of optical character recognition accuracy over other methods in 1996.
Zhange and Malik et al uses the same idea for object recognition. The naive SVM-KNN algorithm they proposed is:
For each query,
\begin{itemize}
\item Compute nearest K neighbor of query point.
\item If K neighbors have all same labels, query is labeled and exit; else computer
pairwise distance between K neighbors.
\item Convert distance matrix into kernel matrix and apply multi-class SVM.
\item Use the above classifier to label the query.
\end{itemize}

This algorithm will be slow. So they proposed another version in which first they apply simple distance measure initially and select some examples. After that use costly distance measure on these selected data points. So now algorithm becomes:
\begin{itemize}
\item Find collection of neighbors using simple distance measure.
\item Compute costly distance measure on these examples and select K from
these examples.
\item Compute pairwise distance between these K points.
\item Apply DAGSVM on the kernel matrix for training.
They reported using this algorithm, achieved approximately 60\% accuracy in the Caltech dataset which some small modification in using some other distance measure.
\end{itemize}
\subsection{Local Ensemble Kernel Learning}
In this paper\cite{lin07}, proposed a model to learn kernels locally. Suppose there are $C$ classes and $S = (x_i, y_i)_{i=1}^l$ where $y_i \in {1, 2, ..., C}$. Suppose there are $M$ feature representation of the data $x_i$. And each feature descriptor $(1 \le r \le M)$ have distance function $d_r$. Kernels $(K_1,K_2,...,K_M)$ are formed using $d_r$ applied in $x^r_i$ . Aim to combine these kernels. But applying kernel learning here will earn the good kernel globally. Here authors tried to combine the kernel locally to each sample. Let the neighborhood of a sample $x_i$ is defined by weighted vector $wi = [w_{i,1}, ..., w_{i,l}]$ where
$$w_{i,j} =(w^1_{i,j} +...+w^M_{i,j})/M $$
$$w^r_{i,j} = exp(-d_r(x_i,x_j)^2) $$
Let the local target kernel of $x_i$ is defined by $G_i(p,q)=w_{i,p} × w_{i,q} × G(p,q)$
Now kernel alignment is done with the $M$ Kernels and local target kernel to get the optimized kernel. Experiments in CalTech-101 dataset gives approximately 61\% accuracy. Most object categorization dataset is manually labelled through crowdsourcing approach \cite{griffin2007caltech,dg6}. Caltech-101 also contains face detection\cite{dg6,dg4} dataset.
\subsection{Improving Local Learning by Exploring the Effects of Ranking}
Here for each labeled sample, the proposed technique learns a local distance function and ranks its neighbors at the same time. For sample I, a distance function to rank its neighbors for improving object classification. Since a closer sample generally implies a higher probability to be included in a k-NN scheme, one would expect the distance function to be learned is affected more by those samples near I, specified by the distance function itself. That is, if we put the samples into an ordered neighbor list according to the measurements by the distance function in an increasing manner, the top portion of the list should be more influential in learning. P-Norm Push in tends to pay more attention to the top portion of a ranked list.
In this method also, there will set of distance function available. And aim is to learn the weighted combination of these distance function for each sample. Weights for the distance function is learned according to maximum margin optimization method. So then given the test sample, training are ranked by the distance function and then according to the rank label is assigned to the test sample. Results in CalTech dataset they showed is around 70\%.
\subsection{Multiple Kernel Learning}
Goal of kernel learning is to learn a kernel which is optimal for the specified task. In multiple kernel learning, suppose there are n given data points $(x_i,y_i)$, where $x_i \in X$ for some input space and $y_i \in {-1,+1}$. Suppose there are m kernel matrices $K_j \in R^{n×n}$. The problem is to learn best linear combination of these kernel. That is to learn $\eta_j$ in
$$\sum_{j=1}^m K_j $$
Bach\cite{lanc04,bach07} et al proposed SMO kind of algorithm for solving these problem along with SVM\cite{ChVaBoMu02,dg3} optimization problem.
Authors applied multiple kernel learning for learning weights for each kernel. Here each kernel is build around each descriptors. Consider there are Nk base descriptors and associated distance function $f_1,f_2,...,f_{Nk}$. Each descriptor is either good at discriminative power or invariance power. Given training set, we need learn weight for these descriptor. Descriptors and distance functions are kernalized to form $K_1,...,K_{Nk}$. Kernel $K_k(x,y)=exp(-\gamma_kf_k(x,y))$.
From the base kernels, optimal kernel $K_{opt} = \sum_k d_kK_k$ where the weight d correspond to trade-off level. Now primal problem,
$$min_{w,d,\xi} w^tw + C1^t\xi + \delta^td $$
subject to,
$$y_i(w^t \phi(x_i) + b) \ge 1 - \xi_i$$
$$\xi_i \ge 0, d \ge 0, Ad \ge p $$$$ \phi^t(x_i)\phi(x_j) = \sum_k d_k\phi_k^t (x_i)\phi_k(x_j)$$
These problem is solved by using dual formulation of the above problem\cite{dg1,dg2,dg5}.
\subsection{SURF: Speeded Up Robust Feature}
Feature descriptors like SIFT\cite{lowe04} takes much computation time for calculating descriptors. So it will be expansive to these operation in devices like mobile phones. SURF features will have less computation time and at the same time has comparable accuracy as SIFT feature descriptors.
\begin{figure}
\centering
\includegraphics[width=3.5in,height=2in]{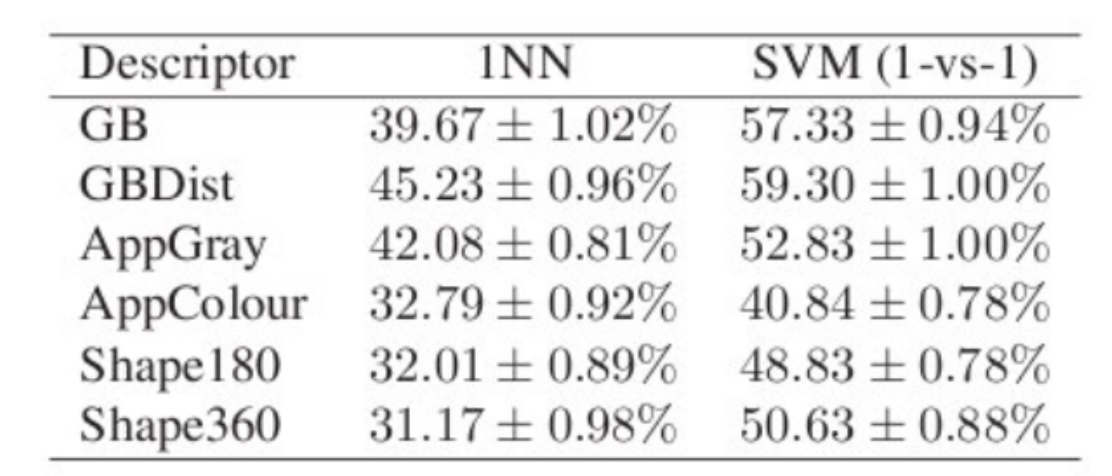}
\end{figure}

\subsection{Adaboost for Combining Descriptors}
Various algorithm for object recognition are experimented. Adaboost\cite{collins2002logistic} SVM mentioned in the results is performed with SVM as set of classifiers. Adaboost mentioned in the result is performed with following setup:
\begin{itemize}
\item Set of classifier for Adaboost is SVM.
\item Each SVM in that set is build on different base kernel $K_i$ mentioned in previous section.
\item Here each $K_i$ are build with descriptors like pyramid histograms of gradient, scale-invariant feature descriptors.
\item Adaboost provides weight on the SVM classifier which is build on each such kernel mentioned above.
\end{itemize}
This AdaBoost is inspired by the Multiple kernel learning work where each kernel is formed with different descriptors. Difference between AdaBoost with SVM(with different descriptors) and Multiple kernel learning is AdaBoost gives weight on the SVM classifier(each SVM with different descriptors in kernel) where in multiple kernel learning gives weights on each kernel in a SVM problem. Following table shows result.
Local learning and AdaBoost with SVM performs better here. But time taken by Local learning is high because of its algorithm. AdaBoost with SVM gives 78\% for 5 object classes which is far less to Multiple kernel learning trained for over 100 object classes. But AdaBoost here is not experimented with many descriptors unlike in multiple kernel learning.
\begin{figure}
\centering
\includegraphics[width=3.5in,height=2in]{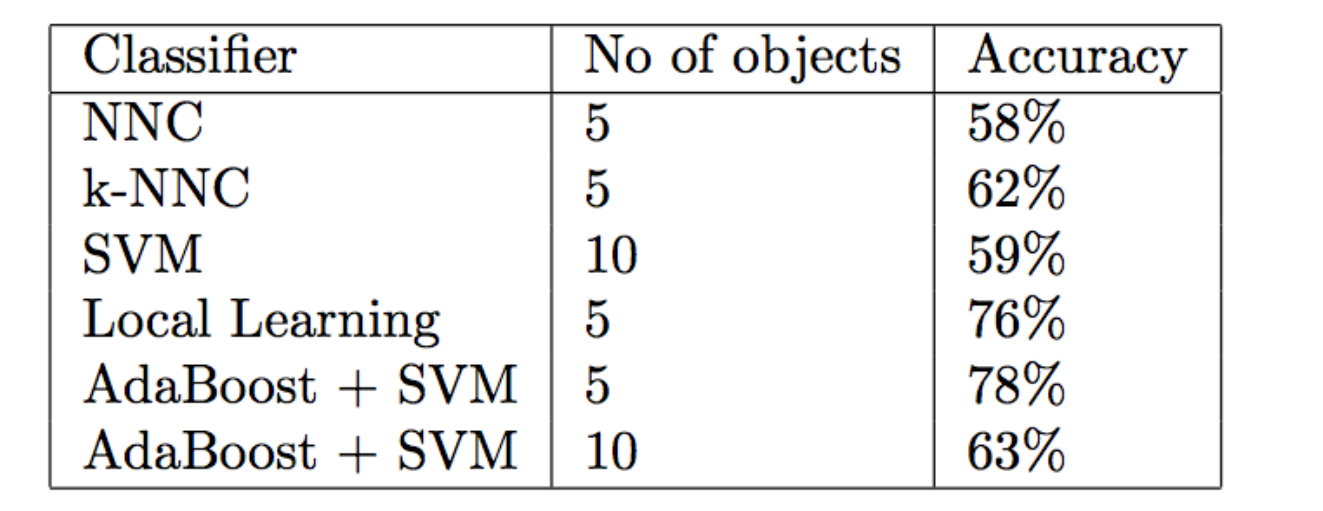}
\end{figure}
\section{Conclusion}
Object categorization problems typically involves using multiple descriptors. This is first attempt to combine these descriptors using Adaboost with SVM as week learners. This study shows that Adaboost with SVM as week learner improves object categorization accuracies by 2.6\% compared to other well known methods. `
{\small
\bibliographystyle{ieee}
\bibliography{egbib}
}

\end{document}